\documentclass[journal,preprint]{IEEEtran}
\usepackage{amsmath,amsfonts}
\usepackage{algpseudocode}
\usepackage{algorithm}
\usepackage{array}
\usepackage[caption=false,font=normalsize,labelfont=sf,textfont=sf]{subfig}
\usepackage{textcomp}
\usepackage{stfloats}
\usepackage{verbatim}
\usepackage{graphicx}
\usepackage{cite}
\usepackage{url}
\usepackage{booktabs}
\usepackage{enumitem}
\usepackage{todonotes}

\usepackage{xcolor}

\definecolor{pastelyellow}{RGB}{255, 245, 180}
\definecolor{darkpastelblue}{RGB}{109, 148, 197}

\usepackage{soul} 

\sethlcolor{pastelyellow}  

\usepackage[final,commandnameprefix=ifneeded]{changes}

\setaddedmarkup{\color{black}\hl{#1}}

\setdeletedmarkup{\textcolor{darkpastelblue}{\sout{#1}}}

\usepackage{hyperref}
\hypersetup{
  pdftitle={UBCL: A Reinforcement Learning Framework for
            Controllable and Diverse Player Behaviors},
  pdfauthor={Atahan Cilan and Atay Ozgovde},
  pdfsubject={IEEE Transactions on Games},
  pdfkeywords={reinforcement learning, player modeling, game AI,
               human-like agents},
  colorlinks=false,
  pdfborder={0 0 0}
}

\begin{document}

\title{UBCL: A Reinforcement Learning Framework for Controllable and Diverse Player Behaviors}

\author{
    Atahan~Cilan
    and~Atay~Özgövde,~\IEEEmembership{Senior Member,~IEEE} 
    \thanks{A. Cilan is with Turkish Aerospace, Istanbul, Turkey, and also with the Department of Computer Engineering, Boğaziçi University, Istanbul, Turkey (e-mail: atahan.cilan@std.bogazici.edu.tr).}
    \thanks{A. Özgövde is an Associate Professor in the Department of Computer Engineering at Boğaziçi University, Istanbul, Turkey 
    (e-mail: ozgovde@bogazici.edu.tr).}
    \thanks{\copyright~2026 IEEE. Personal use of this material is permitted. Permission from IEEE must be obtained for all other uses, in any current or future media, including reprinting/republishing this material for advertising or promotional purposes, creating new collective works, for resale or redistribution to servers or lists, or reuse of any copyrighted component of this work in other works.}
    \thanks{Accepted for publication in IEEE Transactions on Games. DOI: 10.1109/TG.2026.3703355}
}

\markboth{IEEE Transactions on Games}{Cilan and Özgövde: UBCL: A Reinforcement Learning Framework for Controllable and Diverse Player Behaviors}


\maketitle

\begin{abstract}
This paper introduces a reinforcement learning framework that enables controllable and diverse player behaviors without relying on human gameplay data. Existing approaches often require large-scale player trajectories, train separate models for different player types, or provide no direct mapping between interpretable behavioral parameters and the learned policy, limiting their scalability and controllability. We define player behavior in an N-dimensional continuous space and uniformly sample target behavior vectors from a region that encompasses the subset representing real human styles. During training, each agent receives both its current and target behavior vectors as input, and the reward is based on the normalized reduction in distance between them. This allows the policy to learn how actions influence behavioral statistics, enabling smooth control over attributes such as aggressiveness, mobility, and cooperativeness. A single PPO-based multi-agent policy can reproduce new or unseen play styles without retraining. Experiments conducted in a custom multi-player Unity game show that the proposed framework produces significantly greater behavioral diversity than a win-only baseline and \replaced{matches specified behavior vectors across diverse targets, with highest accuracy on score-based dimensions and moderate degradation on cooperation and 
mobility.}{reliably matches specified behavior vectors across diverse targets.} The method offers a scalable solution for automated playtesting, game balancing, human-like behavior simulation, and replacing disconnected players in online games.
\end{abstract}


\section{Introduction}
\begin{figure}[!t]
  \centering
  \includegraphics[width=0.48\textwidth]{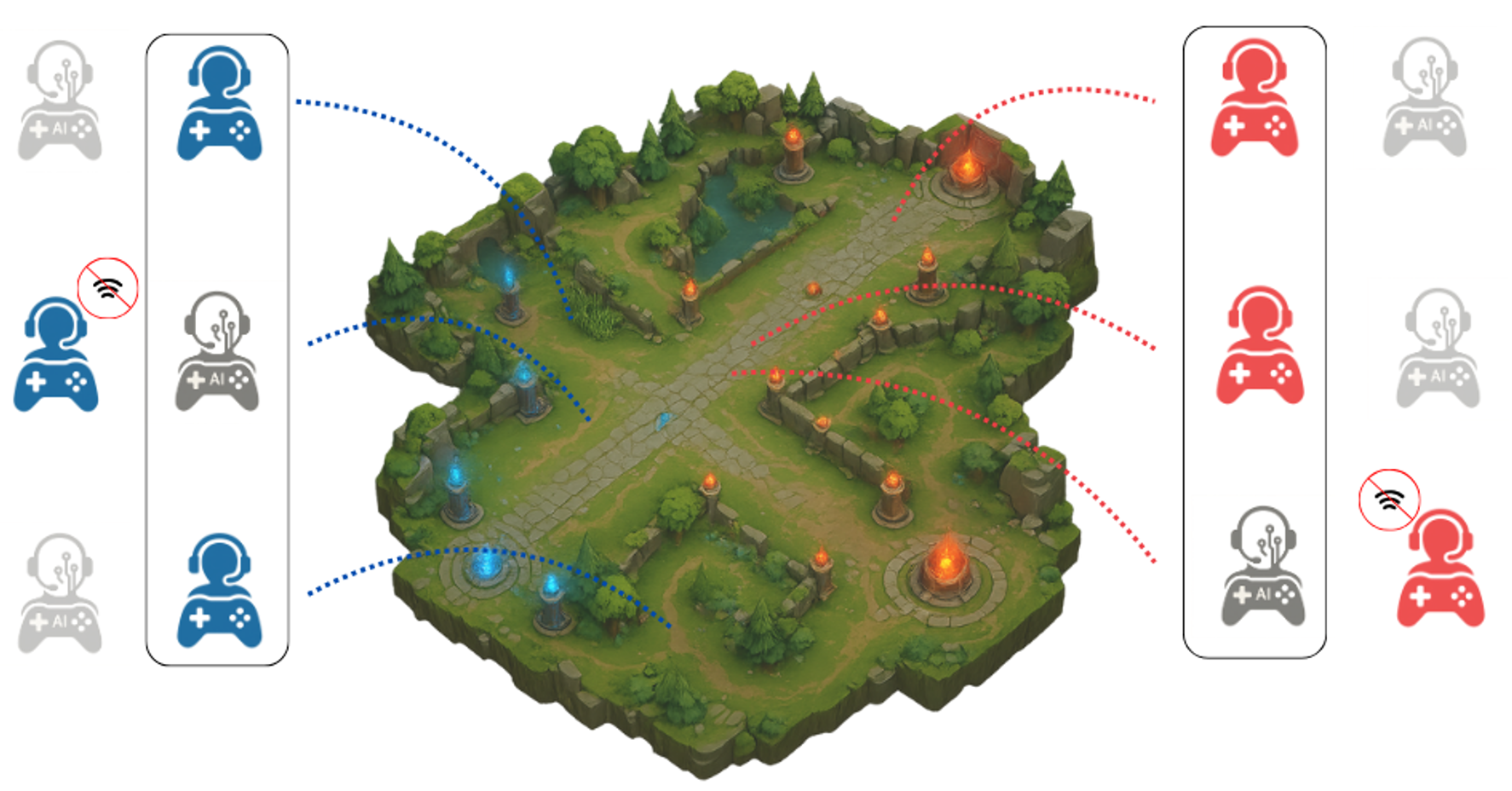}
  \caption{Sample scenario that involves representative models of each player in an online multiplayer game. If a player is disconnected, the AI model corresponding to that player can replace it until the player reconnects.}
  \label{fig:sample_scenario}
\end{figure}

\IEEEPARstart{C}{omputer} games serve as a prominent platform for artificial intelligence (AI) research, owing to their complexity and the vast number of algorithms that have been developed and evaluated within game environments. A central research objective in this domain is to develop AI agents capable of playing games at or beyond human-level performance. Notable studies \cite{dota2_openai, alpha_star, human_level_control, silver2016mastering_go, outperforming_atari} demonstrated impressive results in achieving superhuman game play. Beyond performance-oriented goals, a growing body of research focuses on developing AI agents that emulate human-like behavior. These agents are valuable for applications such as automated play-testing, dynamic game balancing  \cite{arifin2025review_dynamic_game_balancing, jeon2023raidenv_IEEE_dynamic_game_balancing, andrade2006_dynamic_game_balancing}, evaluating game systems like matchmaking \cite{wang2024enmatch_matchmaking_evaluation, ruttgers2024automatic_matchmaking_evaluation}, temporarily replacing disconnected players in online multiplayer games \cite{pfau2021deep_player_modeling}, or any simulation scenario that requires human-like behavior \cite{sreedhar2025simulating_humanlike_simulation, park2023generative_humanlike_simulation}.\added{Recent studies further demonstrate that player behavior is inherently diverse: engagement patterns depend on multimodal contextual factors}~\cite{engagement_estimation}\added{, and behavioral tendencies vary significantly across cultural 
backgrounds}~\cite{cultural_player_behavior}\added{, reinforcing the need 
for player models that can flexibly represent this variety.} Figure \ref{fig:sample_scenario} displays a sample scenario for the usage of human-like player models. To this end, various approaches have been proposed for generating human-like player models with diverse behavioral styles.

Traditional methods typically depend on predefined rules for generating player archetypes, such as killer, defender, explorer, etc. \cite{procedural_persona_mcts, rule_based_game_ai, match3_test}. Rule-based methods significantly simplify the behavior types of the players, can only represent limited player behavior, and often fail to adapt or generalize to new situations \cite{procedural_persona_mcts, rule_based_game_ai, match3_test}. Beyond rule-based approaches, deep learning based methods are frequently studied for generating human-like player models. Imitation learning approaches \cite{human_like_3rd_person_shooter, ahlberg2023generatingpersonasgamesmultimodal, imitiation_of_boardgame_players, dung_n_replicants_2, behavior_modelling_rpg, synt_user_gen, pfau2021deep_player_modeling, Le_Pelletier_AILAD} use human-generated trajectories as a reference for training the player models. The success of imitation learning based methods heavily depends on the quality of human-generated data and often struggles to adapt to unseen game states\cite{human_like_3rd_person_shooter, 
ahlberg2023generatingpersonasgamesmultimodal, 
imitiation_of_boardgame_players, pfau2021deep_player_modeling}. 

\IEEEpubidadjcol

Similarly, a growing literature of reinforcement learning (RL) and inverse reinforcement learning (IRL) approaches exists for generating player models with distinct behaviors. IRL based methods use human-generated data to learn a proper reward function, then apply RL to learn diverse policies. Several studies \cite{football_agents_by_irl, beyond_wl_motivation} successfully achieve player models with different play styles. However, large collections of human game-play data are needed to capture different play strategies, which significantly affect the performance. Gathering such data is costly and time-consuming.

On the other hand, RL-based methods frequently use extensive reward shaping for generating different play styles. Methods like \cite{Le_Pelletier_CARI, curiosity_driven_rl, autogametesting_rl_2020, schnabl2024mario_bros, rl_4_beliavable_bots} successfully generate various play styles by applying reward shaping to diversify the trained model's behavior. Despite the success of reward shaping in generating behaviorally rich player models, these methods cannot be used to directly represent a known human player, as they require knowing the exact reward coefficients for the player. As suggested in the study \cite{Le_Pelletier_CARI}, a model that predicts reward coefficients from the game metadata of human players is needed for this task. In other words, a mapping between human players and the trained player models is needed for meaningful representation. 

Therefore, existing methods fail to generate \replaced{diverse}{rich, human-like} player behaviors using a single model while simultaneously enabling a direct mapping between a human player and the trained model, without relying on human-generated data. With this motivation, we propose an RL framework based on the uniform sampling of desired behaviors for conditioning the policy to learn rich and directly usable player models, which we call Uniform Behavior Conditioned Learning. Our Uniform Behavior Conditioned Learning (UBCL) framework enables agents to acquire and display diverse game-play behaviors without requiring human demonstrations. The framework allows agents to imitate a wide range of play styles, each encoded as a compact vector, solely through environment interaction. Furthermore, it supports direct mapping from human player data to agent inputs, enabling the representation of specific human players using the trained agent without any further training.

UBCL framework utilizes the vector representation of an expected play style, which we call the behavior vector, to determine the agent's behavior in a game. Important behavioral features like aggressiveness and cooperativeness are converted to numerical values and embedded into the behavior vector. Each dimension in the behavior vector represents a certain behavioral feature in the range between 0 and 1, where 0 means minimum expression of this behavioral feature and 1 means maximal expression. For example, an aggressiveness value of 0 may correspond to a complete lack of motivation to engage in lethal actions against opponents, whereas a value of 1 may indicate a willingness to eliminate opponents by any means necessary. During the training, behavior vectors are sampled from a uniform distribution for each player per game and assigned as the target behavior vector. At each time step of the game, agents observe the game states, the target behavior vector, and their current behavior vector. The current behavior vector conveys information regarding the agents’ present behavioral state. The reward function is defined by the normalized change in the Euclidean distance between the current and target behavior vectors, which is formulated to incentivize actions that align the agents’ current behavior vector with the target behavior vector, as well as to maintain the existing behavioral state when alignment has already been achieved. Uniform sampling of the target behavior vectors enables agents to acquire the full spectrum of possible behavior types within the multidimensional space from which these vectors are derived.

To test our framework, we designed a custom multi-agent game environment in Unity and utilized the Unity ML-Agents toolkit \cite{unity_ml} for training the policy. Game environment models a 2v2 team-based competition in a grid arena, where players collect various objects for points and engage in combat. Agent behavior is guided by a six-dimensional target behavior vector capturing aspects such as cooperativeness, competitiveness, aggressiveness, mobility, and risk-taking. Observation space of the agents includes the target behavior vector, current behavior vector, map perception data, and other game-state features.

Using this setup, a policy was trained with Proximal Policy Optimization (PPO) \cite{ppo_paper} to produce consistent and distinguishable behaviors aligned with target behavior vectors. In contrast to prior studies that rely on reward coefficient diversification or utility function adjustments \cite{procedural_persona_mcts, Le_Pelletier_CARI, schnabl2024mario_bros, rl_4_beliavable_bots}, the UBCL framework leverages actual game metrics in the reward function, \replaced{enabling approximate reproduction of desired human-like behaviors by using game-play data, with accuracy varying across behavioral dimensions.}{enabling direct reproduction of desired human-like behaviors by using game-play data.} Furthermore, our UBCL framework utilizes information obtained from non-feasible behavior targets in addition to the feasible ones, thereby eliminating the need for human data, as its primary purpose is to provide meaningful play style references for the policy to learn. Results are presented through radar chart visualizations, which demonstrate the agent’s ability to adapt to diverse stylistic objectives. In addition, we trained a win-only policy \added{and a CARI policy}\cite{Le_Pelletier_CARI} to validate the behavioral richness of the policy trained with the UBCL framework. We provided a 2D visualization by using dimensionality reduction for comparing behavioral style coverage of the \replaced{baselines}{win-only agent} and the policy trained with the UBCL framework.

In the following section, existing methods for human-like player behavior modeling are reviewed. The UBCL framework is described in detail in the Methodology section. Implementation details are outlined in the Experimental Setup section, \replaced{experimental findings are presented in the Results section, and their implications are examined in the Discussion section.}{the experimental findings are presented in the Results and Discussion section.}

\section{Related Work}
Research on human-like player behavior modeling has evolved along several complementary directions, ranging from supervised imitation of human trajectories to reinforcement learning methods that synthesize plausible or stylistically controlled play-styles. This section reviews the main approaches and their limitations in terms of human-generated data dependence, scalability, and controllability. Here, \emph{scalability} refers to how easily a trained model can be adapted to new or unseen play-styles, while \emph{controllability} describes the extent to which specific or known player behaviors can be intentionally reproduced by the model.

\subsection{Imitation Learning with Human Demonstrations}

A significant number of studies on player behavior modeling rely on imitation learning to directly reproduce trajectories from human game-play data. Farhang \emph{et al.}~\cite{human_like_3rd_person_shooter} trained a causal transformer on third-person shooter sessions, conditioning on player identity tokens to reproduce distinct aiming or movement styles within a single network. Ahlberg \emph{et al.}~\cite{ahlberg2023generatingpersonasgamesmultimodal} introduced MultiGAIL, which employs multiple discriminators to interpolate smoothly between player personas during runtime. Pan \emph{et al.}~\cite{imitiation_of_boardgame_players} proposed a meta-learning approach that rapidly personalizes a generic model to new board-game players with few samples, while Pfau \emph{et al.}~\cite{dung_n_replicants_2} developed player-specific replicant agents for balancing role-playing games using real user data. Chapa~Mata \emph{et al.}~\cite{synt_user_gen} combined transformers with diffusion models to mirror play styles of known players in an action-adventure game.  
 
Collectively, these approaches demonstrate high fidelity to observed human trajectories but share a key limitation: they depend heavily on large-scale demonstration datasets and struggle to generalize to unseen or intentionally designed behaviors \cite{human_like_3rd_person_shooter, ahlberg2023generatingpersonasgamesmultimodal, imitiation_of_boardgame_players, pfau2021deep_player_modeling}. As a result, imitation learning–based methods provide limited scalability and controllability, despite their strong performance when large-scale human demonstration data are available.

\subsection{Inverse Reinforcement Learning and Preference Modeling}

Several studies employ inverse reinforcement learning (IRL) to learn reward functions that explain human play. Wakabayashi \emph{et al.}~\cite{football_agents_by_irl} applied IRL to soccer simulations, yielding more contextually appropriate movement and decision patterns compared to hand-crafted rewards. Wang \emph{et al.}~\cite{beyond_wl_motivation} extended this idea through multi-dimensional IRL to capture motivational factors in MMORPGs, highlighting that different player groups exhibit distinct reward structures.  

While these works align agents with implicit human preferences and improve believability, they remain difficult to control once a reward function is learned. Generating new or targeted play styles often requires retraining from scratch, limiting flexibility in design-time personalization. Similar to imitation learning methods, IRL-based approaches also face scalability and controllability limitations.

\subsection{Reinforcement Learning for Procedural and Human-Like Play}
 
More recent research has explored RL frameworks that generate human-like and stylistically varied behaviors without relying entirely on demonstration data. Ho \emph{et al.}~\cite{ho2023humanlikerltamingnonnaturalistic} introduced Adaptive Behavioral Costs (ABC-RL), which regularizes RL agents to suppress unnatural motion artifacts, improving perceived realism of AI players in 3D games. Similarly, Glavin and Madden~\cite{rl_in_fps_bots} and Arzate~Cruz and Ramírez-Uresti \cite{rl_4_beliavable_bots} pioneered the integration of RL into first-person shooters to enhance believability. The primary goal of these studies is to enhance the human-likeness of trained player models. They try to diversify the learned behaviors to further improve human-likeness. Nonetheless, they lack specific methods for directing the learned policy to demonstrate a particular play style.

An important application of player models exhibiting diverse play styles is automated playtesting, and several RL–based approaches have been proposed to address this objective. Bergdahl et al.~\cite{autogametesting_rl_2020}, Gordillo et al.~\cite{curiosity_driven_rl}, and Schnabl~\cite{schnabl2024mario_bros} demonstrated that deep RL agents can autonomously explore game environments for level testing. However, such agents often converge to repetitive and non-human optimal behaviors. Although these studies aim to enhance behavioral diversity to improve test coverage, they lack mechanisms to control or direct the learned policy toward specific play styles. 

Le~Pelletier~de~Woillemont \emph{et al.}~\cite{Le_Pelletier_CARI} previously introduced CARI, an RL framework designed to produce diverse behavioral profiles without relying on human demonstration data. In this method, agent behaviors are shaped by varying reward coefficients, enabling the generation of multiple play styles within a single environment. However, since there is no explicit mapping between these reward coefficients and human behavioral data, reproducing a specific human play style remains challenging. To address this limitation, the authors later proposed CARMI, an RL framework that utilizes human data as a reference and generates play styles consistent with human data~\cite{Le_Pelletier_CARMI}.

Recently, Sun~\emph{et~al.}~\cite{enhancing_ai_bot_strength} introduced a deep reinforcement learning framework for adversarial games that enhances AI-bot strength and strategy diversity without relying on human data. While their approach demonstrates impressive performance in generating competitive and varied strategies, it is primarily suitable for bot design rather than personalized player emulation, as it lacks explicit controllability over behavioral traits. 

Overall, reinforcement learning approaches advance scalability and autonomy but often lack explicit mechanisms to steer agents toward specific human-like personas. Bridging the gap between controllability and diversity remains an open challenge in player behavior modeling.

\subsection{Theoretical Positioning}

\added{
The approaches reviewed above can be situated within 
broader reinforcement learning paradigms. Multi-Objective 
RL (MORL) methods search for Pareto-optimal trade-offs 
among competing reward signals}~\cite{hayes2022morl}\added{, 
while Quality-Diversity (QD) algorithms such as 
MAP-Elites}~\cite{mouret2015mapelites}\added{ maintain an archive 
of diverse, high-performing solutions without accepting 
user-specified targets. Unsupervised skill discovery 
methods such as DIAYN}~\cite{eysenbach2019diayn}\added{ learn 
distinguishable skills by maximizing mutual information 
between latent skill variables and visited states, but 
likewise do not support user-specified behavioral targets. 
UBCL is most accurately described as a form of 
Goal-Conditioned RL 
(GCRL)}~\cite{liu2022gcrl}\added{: the agent receives a target 
point in a bounded, interpretable behavior space and is 
rewarded for reducing its distance to that point (Eq.~1). 
The behavior vector serves a dual role depending on the 
dimension type: ratio-based dimensions (e.g., score 
composition) act as style constraints that must be tracked 
throughout the episode, whereas cumulative dimensions 
(e.g., total score attainment) behave as end-state goals 
that can only be satisfied asymptotically. In both cases, 
the underlying mechanism is the same---minimization of 
distance to a user-specified point---which distinguishes 
UBCL from MORL (no Pareto search), QD (user specifies the 
target rather than an archive), DIAYN (no user-specified 
target), and reward-conditioned methods such as 
CARI}~\cite{Le_Pelletier_CARI}\added{ (point targeting rather 
than directional control via reward weights). A detailed comparison is provided in Section~VI.
}

\section{Methodology}

Despite the increasing focus on player modeling in recent studies, most existing methods either rely on extensive human demonstrations or lack fine-grained controllability over behavioral style.

Imitation and IRL-based approaches tend to reproduce or infer from existing human data, while pure RL frameworks prioritize performance or diversity without ensuring interpretable and direct persona control. In contrast, our \emph{Uniform Behavior Conditioned Learning (UBCL)} framework aims to bridge this gap by conditioning the policy on explicitly defined behavioral targets rather than implicit demonstrations or inferred rewards. This design enables flexible and data-efficient generation of distinct play-styles without retraining, providing a unified mechanism for controllable and human-like behavior synthesis.

UBCL framework trains a single model that can be used to represent the desired player's behavior with known game metadata, which can be obtained based on a limited amount of play sessions - as low as a single game play.

In each episode, every agent samples a target behavior style from a uniform distribution defined over the full behavior space and attempts to match the corresponding behavioral statistics. Although real players typically occupy a non-uniform and more limited subset of this space, covering the entire domain enables the model to explore and learn how actions affect a broad range of behavioral outcomes. Consequently, the agent acquires a policy capable of achieving high rewards whenever a valid target behavior is sampled, including those that fall within the subset of human players.

\begin{figure}[!t]
\centering
\includegraphics[width=0.48\textwidth]{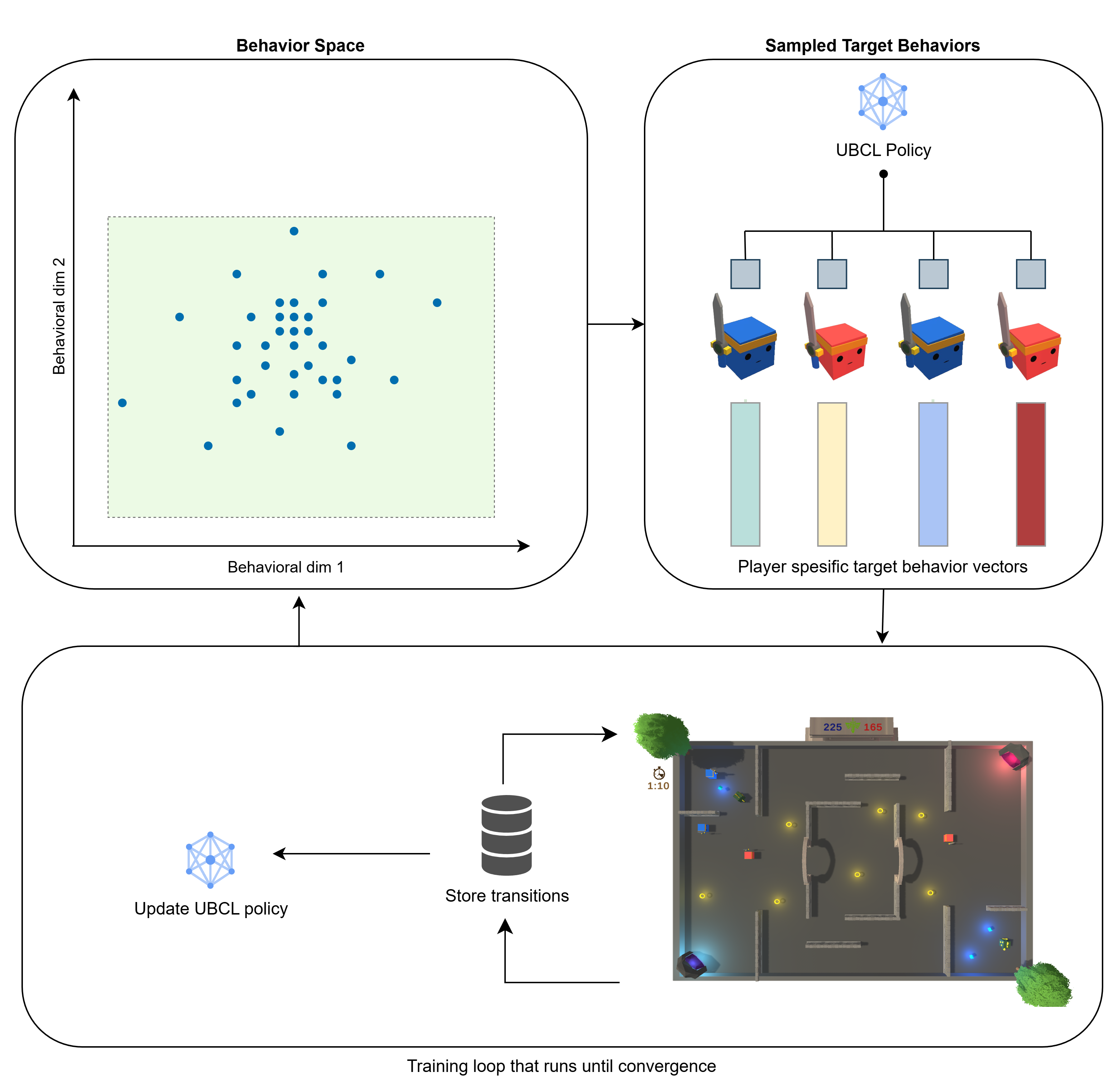}
\caption{
Overview of the UBCL approach. The behavior space in the upper-left illustrates the possible play styles for two behavioral parameters. Blue points denote hypothetical play styles that represent real players, highlighting that the target behaviors sampled during learning span a broader region than the actual player styles. For each episode, distinct play-style targets are sampled for every player, and the resulting gameplay data are used to update the UBCL policy.
}
\label{fig:ubcl_framework}
\end{figure}

Figure \ref{fig:ubcl_framework} presents an overview of the UBCL framework. The blue points in the behavior space illustrate hypothetical play styles that real players might exhibit but are not directly observable due to the absence of human gameplay data.

In this light, our method supports creating a continuum of play-styles, including human-like play styles that can be used for player modeling and non-human-like play styles that can be used for comprehensive play testing. The following sections describe the definition of behavior vectors and the training process.

\subsection{Behavior Vectors}
In our study to facilitate a single RL model to cover a wide spectrum of possible user profiles, we introduce the concept of behavior vector. The behavior vector is a compact description of certain behavioral tendencies that summarize the player’s play style. Each element in this vector is expected to represent a desired behavior dimension with a numeric value.

Although defining vectoral dimensions may vary across different games, depending on the genre, choices become apparent. For example, the “kill/assist ratio” of a player in a MOBA or FPS game is a natural candidate for a dimension of the behavior vector. This ratio reflects the aggressiveness or cooperativeness of the player. Similarly, the “weapon preference ratio”, which indicates the player’s tendency to select ranged weapons over close combat, captures a crucial aspect. Although it is possible to use any custom internal metric that reflects a certain user behavior, it is imperative that these values should be calculable at any step of the game. Sample behavioral parameters are presented in Table I for various game genres.

\begin{table}[h]
\centering
\caption{Example Behavioral Parameters Across Game Genres}
\label{tab:behavior_dimensions}
\begin{tabular}{l p{4.2cm}}
\hline
\textbf{Game Genre} & \textbf{Example Behavioral Parameters} \\
\hline
FPS (Shooter) &
Kill/Assist Ratio; Aggression Level; Map Control Efficiency; Weapon Preference Patterns \\

MOBA &
Lane Pressure; Farming Efficiency; Team-fight Positioning; Role Fidelity \\

RPG / Action-Adventure &
Exploration Depth; Loot Collection Style; Dialogue Choice Patterns; Combat Aggressiveness \\

Puzzle / Strategy &
Problem-Solving Speed; Move Optimality; Pattern Recognition Behavior \\

Sports Game (e.g., FIFA, NBA) &
Pass Accuracy; Shot Selection Quality; Offensiveness–Defensiveness Index \\

Platformer & Exploration vs. Speedrun Tendency; Enemy Avoidance Preference; Collectible Completion Rate \\
\hline
\end{tabular}
\end{table}

In the UBCL framework, behavior vectors are the basic building blocks for conditioning the RL policy. The current behavior vector and target behavior vectors are used to reach a certain behavioral state and keep this behavioral state during the game. Section IV describes an example case on how to define the behavior vectors for a real-world multiplayer game scenario.

\subsection{Reward Definition}

At each step, agents receive a reward based on the normalized change in distance between their current behavior vector and target behavior vector. Specifically, the reward is calculated as:

\begin{equation}
\label{eq_reward}
r_t = \lambda \cdot \frac{(\|b_{t-1} - b_{\text{target}}\|_2 - \|b_t - b_{\text{target}}\|_2)}{\| b_{\text{target}}\|_2 } 
\end{equation}

where $b_t$ is the current behavior vector, $b_{\text{target}}$ is the fixed target behavior vector for the episode, and $\lambda$ is a scaling coefficient. Positive reward is given when the agent’s actions reduce the Euclidean distance to the target behavior vector; negative reward is given otherwise. After reaching the desired behavioral state, which means obtaining the same or very close current behavior and target behavior vectors, the agent also tries to keep this state since any action that is not aligned with the expected behavior drives the current behavior vector farther from the target, and the agent receives negative rewards for that action. This structure encourages the agent to learn diverse styles and align its episodic behavior with varied behavioral objectives, without any imitation from human game-play data. One important detail about the reward definition is normalizing the reward with the magnitude of the target behavior vector. Without normalization, the max return (sum of rewards) for an episode equals the magnitude of the target behavior vector. However, each player's behavior can be represented with a target vector with different magnitudes. As an example, a player can be quite offensive and cooperative at the same time, meaning we would represent this player with a value of 1 for these two behavior parameters. Similarly, we would represent less offensive and cooperative players with smaller values for these parameters, which makes the magnitude of the target behavior vector smaller. As a result, the maximum achievable return for the second case is smaller than the first one, and this pushes the agent to learn edge behaviors instead of the average behaviors to maximize mean return. To avoid this, we normalized the reward with the magnitude of the target behavior vector. Intuitively, this normalization gives information about the importance of certain actions for certain behaviors. For a player with an offensive nature, killing a single player has less significance since it is common behavior for that behavior type. However, for a more peaceful player, even killing a single player is important and is avoided throughout the game. Applied normalization introduces these types of behavioral traits to the algorithm.  Furthermore, the max return is equal to the $\lambda$ due to normalization, and this makes the training reward more interpretable.

\subsection{Training Process}

The target of the training process is not solely to win the game or collect more points, but rather to shape the agent’s episodic behavior to match the predefined behavior vector. 
Observation space of the agents includes similar game states to a real player, which can be the game map, current scores, game statistics, player positions, etc. On top of game-specific observations, agents also observe the current behavior vector, which is updated at every step, and the target behavior vector, which is sampled at the beginning of each episode.

In each training episode, every agent uniformly samples a target behavior vector from a predefined continuous behavior space. The design of this space should be guided by domain-specific data to avoid sampling unrealistic or infeasible behavior targets. However, in most games, this is not feasible, as some players tend to push the boundaries of the gameplay. Moreover, certain target values may be inherently unrealistic — for instance, expecting all players in a team to achieve only assist scores without any kills. In practice, at least one player will inevitably obtain kill scores. Previous studies, such as \cite{ahlberg2023generatingpersonasgamesmultimodal, Le_Pelletier_CARMI, pfau2021deep_player_modeling, barthet2022behave_and_experience}, address this issue by utilizing human gameplay data to define feasible behavior targets. On the other hand, our UBCL method prioritizes learning the actions that drive the current behavior vector towards the target behavior vector. Therefore, when an unreachable target behavior vector is sampled, the agent still takes positive rewards from actions that change the behavior vector in the direction of the target behavior vector, and takes negative rewards for the opposite. Therefore, unrealistic behavior targets are also a useful information source in our method. Algorithm 1 describes the training procedure in detail. UBCL method is agnostic to the selected RL algorithm. Both on-policy methods such as PPO \cite{ppo_paper} and TRPO \cite{trpo_paper}, and off-policy methods such as SAC \cite{sac_paper} and DDPG \cite{ddpg_paper}, can be used. However, off-policy algorithms require substantially more memory when the observation space includes visual inputs. Therefore, we utilized the PPO for learning the policy due to its stability, performance, and lower memory requirement.

\begin{algorithm}[ht!]
\caption{Training with Behavior-Conditioned Rewards}
\begin{algorithmic}[1]
\State Initialize policy network $\pi_\theta$, value network $V_\phi$
\State Initialize replay buffer $\mathcal{D}$
\For{each training episode}
    \For{each agent $i$ in environment}
        \State Sample random behavior target vector
        \State $b^i_{\text{target}} \sim \mathcal{U}[0,1]^6$
        \State Reset agent state and set current behavior vector
        \State $b^i_0 \gets \vec{0}$
    \EndFor
    \For{each timestep $t = 1$ to $T$}
        \For{each agent $i$}
            \State Observe state $s^i_t$ (grid, vector obs, $b^i_{t-1}$, $b^i_{\text{target}}$)
            \State Sample action $a^i_t \sim \pi_\theta(a | s^i_t)$
            \State Apply action $a^i_t$
            \State Receive environment feedback
            \State Update behavior vector $b^i_t$ with game statistics
            \State Compute reward:
            \[
            r^i_t = \lambda \cdot \frac{(\|b_{t-1} - b_{\text{target}}\|_2 - \|b_t - b_{\text{target}}\|_2)}{\| b_{\text{target}}\|_2 } 
            \]
            \State Store transition $(s^i_t, a^i_t, r^i_t, s^i_{t+1})$ in $\mathcal{D}$
        \EndFor
    \EndFor
    \State Use transitions from $\mathcal{D}$ to update $\theta, \phi$ using PPO
    \State Clear $\mathcal{D}$
\EndFor
\end{algorithmic}
\end{algorithm}

\section{Experimental Setup}
\label{exp_setup}

This section presents the experimental details. The following subsections describe the design of the game environment, the construction of the behavior vector, and the configuration of the PPO algorithm, including its network inputs and training hyperparameters.

\subsection{Environment Design}
\label{env_design}

\begin{figure*}[!h]
\centering
\includegraphics[width=0.85\textwidth]{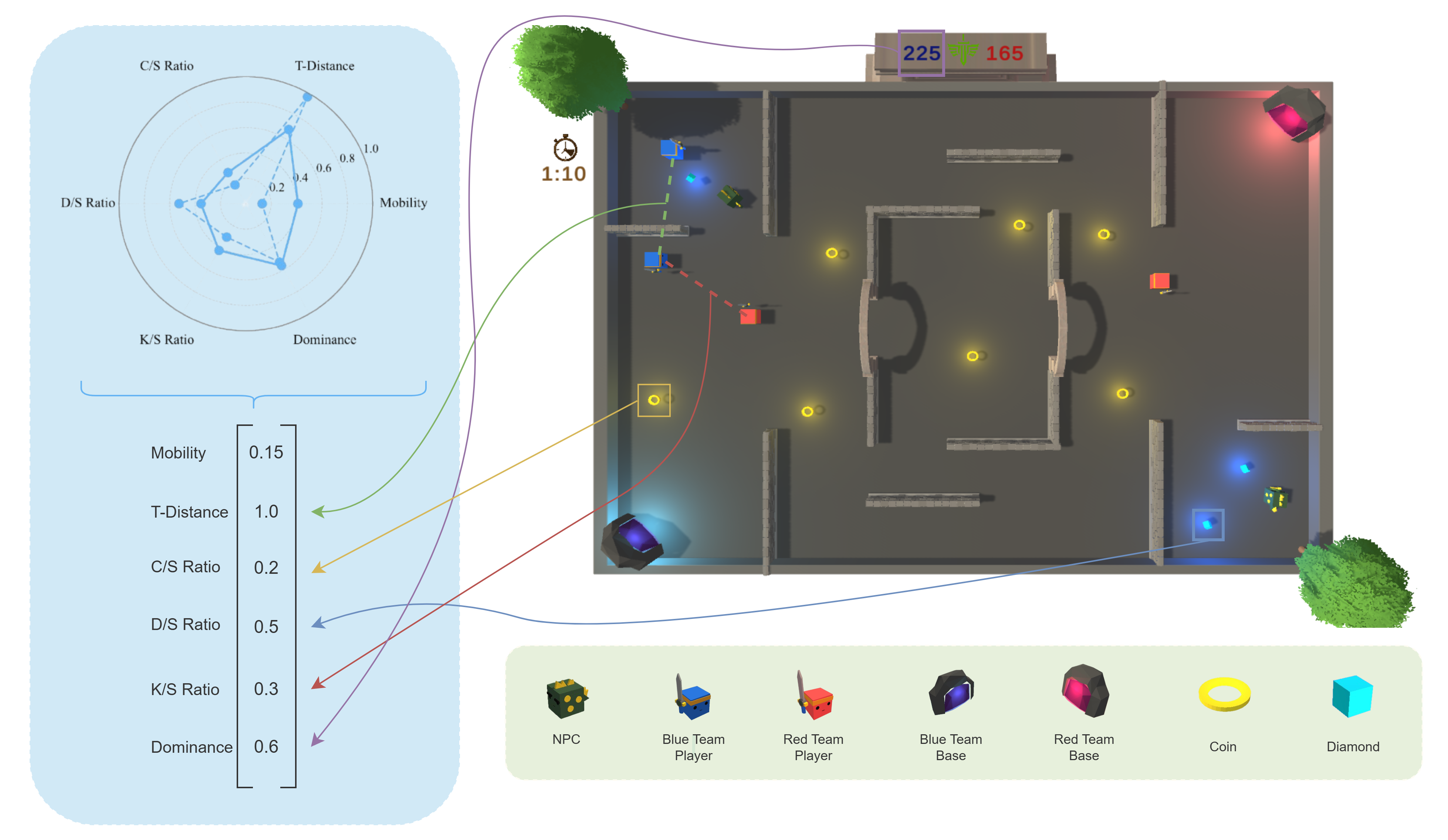}
\caption{In-game screenshot of the custom 2v2 Unity environment. The arena contains walls, coins, diamonds, and two opposing teams. Player behavior is characterized by the parameters listed on the left, each representing a metadata feature derived from distinct game components. Colored arrows indicate the game objects or states associated with each behavioral parameter.}
\label{fig:env}
\end{figure*}

The game environment was developed using the Unity game engine, incorporating a custom-built 2v2 multiplayer game scenario. The decision to create a custom environment was driven by the absence of open source alternatives that met two critical requirements: (1) support for complex, multi-agent competitive and cooperative game dynamics and (2) compatibility with low-resource training setups for iterative experimentation.

The game is played on a discrete grid-based map, where players compete to maximize their score by collecting objects or defeating opponents in a limited time. Players have six discrete actions, which are moving left, right, up, or down, attacking, and waiting. The score system comprises three primary sources: coins (low-risk collectibles that spawn throughout the map), diamonds (high-risk collectibles that spawn close to NPCs that move randomly and kill the player instantly if they touch), and kills (eliminations of opponents). The game includes spatial and temporal randomness in item spawn positions and player locations to prevent overfitting. Coins are generated more frequently than diamonds, and their score contribution is also significantly less. The attack generates an explosion within a range in front of the attacker, and the first object encountered takes the damage. Attacking a teammate is not allowed. In addition, walls and other obstacles can impede the attack. Additionally, attacking collectibles removes them from the map, creating a game mechanic that prevents the opposing team from collecting them.

There are two teams (blue and red) in the game, and each team consists of two players. In our experimental setup, each player is controlled by an RL agent during training. Outside of training, a player can be controlled by a human player or an RL agent. A human player perceives the game through a top-down view of the game map. Human users see the map in continuous form (no grid display), but all actions are executed discretely. A top-down view of the game map is presented in Figure \ref{fig:env} and game elements that are relevant to the construction of the behavior vector are indicated.

RL agents perceive the game map using Unity ML-Agents' grid sensor modules \cite{unity_ml}, which enable spatial encoding of objects on the map, including the controlled player, teammates, enemies, walls, and collectible objects. We generate an eight-channel image-like tensor with a size of (8, 80, 80) by using these sensors. The first dimension encodes the object type (wall, coin, diamond, etc.), and the other two dimensions encode the position of the same type of objects in the map. Additional game states are provided as a vector input, including team ID, remaining time, health, and orientation for each player. Grid observations are inherently normalized, since they encode the spatial information of different object types in binary form. Other vector inputs are normalized to the range (0, 1) by min-max scaling. 

\subsection{Behavioral Parameters}

In addition to the game state observations, we provide the current behavior vector and target behavior vector as inputs to the RL agents, as explained in the Methodology section. For our experiments, we selected six metrics to describe the general behavior of a player. The basic motivation behind the selection of these specific metrics is to describe the player's intentions with clearly understandable and easily calculable metrics. Explanation and formula for each metric are provided below; 

\begin{description}[style=nextline, labelwidth=2cm, labelindent=0cm, leftmargin=!]

  \item[\textbf{Mobility:}] Describes the player's motivation for wandering the map. Calculated by the percentage of visited cells in the grid with the formula \[ b_6 = \frac{N_{\text{visited}}}{N_{\text{total}}} \times 100\] where \( N_{\text{visited}} \) denotes the number of unique grid cells that have been visited by the player, and \( N_{\text{total}} \) is the total visitable number of cells in the grid. 

  \item[\textbf{T-Distance:}] Describes the player's tendency to follow the teammate or act independently. Calculated by \[ b_5=\frac{\frac{1}{N}\sum_{t=1}^N \|x_{t} - x^{\text{teammate}}_t\|_2}
  {d_{max}}   \]
  where  \( d_{max} \) is the maximum distance value possible for the game map. 
  
  \item[\textbf{C/S ratio:}] Proportion of score obtained from coins to total score. Represents the player's tendency to exploit safe resources without engaging opponents or NPCs. Calculated by \[ b_1=\frac{s_c}{s_c + s_d + s_k}\] 
  
  \item[\textbf{D/S ratio:}] Proportion of score obtained from diamonds to total score. Represents the player's tendency to utilize scarce, high-risk, and more valuable resources. Calculated by \[ b_2=\frac{s_d}{s_c + s_d + s_k}\]
  
  \item[\textbf{K/S ratio:}] Proportion of score obtained from opponent eliminations to total score. Describes the aggressiveness of the player. Calculated by \[ b_3=\frac{s_k}{s_c + s_d + s_k}\]
  
  \item[\textbf{Dominance:}] Total score pursuit tendency. Calculated by dividing the total score by a theoretical score limit, where the formula is \[ b_4=\frac{s_c + s_d + s_k}{s_{max}}\]
  
\end{description}

Each metric is normalized to the range [0, 1], which is also included in the equations above. All of these metrics are game metadata and can be calculated for a human player at any time step of the game. Therefore, we can easily calculate these behavior metrics for a human player at the end of the game and achieve a summary of the player's behavior for that game, which is described by these 6 values. However, player motivation and performance can change through different game sessions; therefore, we recommend collecting more than one behavior vector for a player and using the mean of these behavior vectors to represent desired player behavior.

\subsection{Training and Network Architecture}
\label{net_arch}

\begin{figure*}[!t]
\centering
\includegraphics[width=0.98\textwidth]{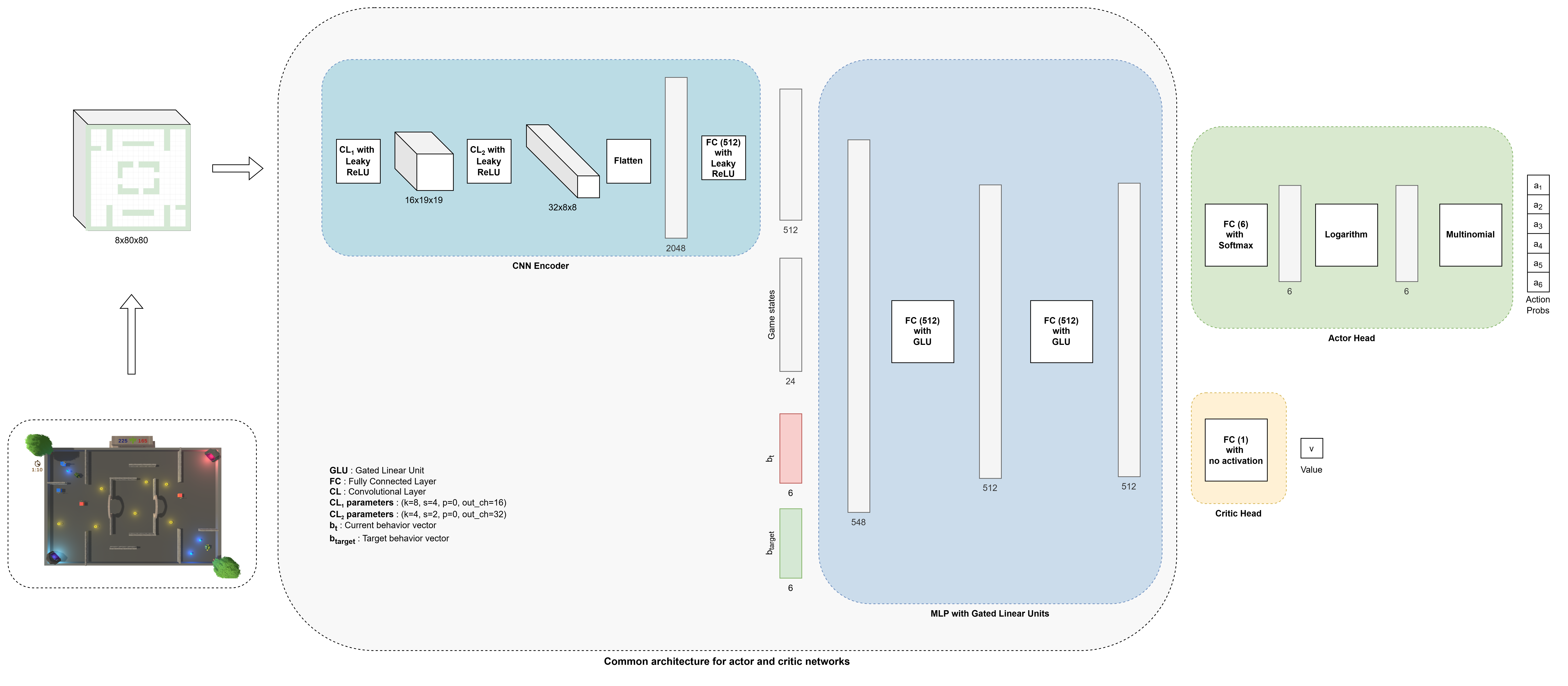}  
\caption{Overview of the agent network architecture. The model processes three observation types: game map, game states, and behavior vectors. The game map is converted into a grid representation and encoded separately using convolutional layers with Leaky ReLU activation. The resulting features are concatenated with the game states and behavior vectors to form a shared representation, which is then passed through fully connected layers and divided into actor and critic branches for PPO training.}
\label{fig:network}
\end{figure*}

To model diverse player behavior, for each episode, every agent is conditioned on a randomly selected six-dimensional \textit{target behavior vector}, which defines the agent’s objective in terms of behavioral style (e.g., aggressive, cooperative, score-maximizing) as it was explained in the Methodology section. Considering that no human game-play data is used during the training part, we introduced some domain knowledge to the sampling of target behavior vectors. For sampling the target values for \textit{C/S ratio}, \textit{D/S ratio}, \textit{K/S ratio}, we utilized the knowledge that their summation should be 1 due to the definition of these values. \textit{Dominance} is sampled from the full range since achieving no score, and the max score is possible. On the other hand,  \textit{T-Distance} and \textit{Mobility} are sampled from a limited range, considering that a zero mean distance between teammates and zero mobility is unrealistic. Therefore, the sampling interval is redefined for this specific game environment instead of directly using $\mathcal{U}[0,1]^6$ (check Algorithm 1) and presented in Table \ref{tab:target_distribution} for each behavioral parameter.

The policy and value networks follow a standard actor-critic design, as employed in Unity ML-Agents’ implementation of Proximal Policy Optimization (PPO). The agent receives three types of observations from the game environment:

\begin{description}[style=nextline, labelwidth=2cm, labelindent=0cm, leftmargin=!]
  \item[\textbf{Game map:}] A multi-channel 2D tensor representing object locations and types in the map with a size of (8,  80, 80) is observed by the agents. CNN layers are applied to this tensor for extracting spatial features. Then, resulting features are concatenated with other input features, as it is depicted in Figure \ref{fig:network}.  
  
  \item[\textbf{Game state:}] General game state parameters, such as health and orientation of each player, team ID, and remaining time, are included in this vector observation, which is concatenated with extracted CNN features and behavior vectors.
  
  \item[\textbf{$b_{t}, b_{target}$:}] Current behavior vector and the episode-specific target behavior vector are provided as input to the policy and value networks. The current behavior vector expresses the achieved behavioral state at any game step, and the agent uses the error between the current behavior vector and the target behavior vector to select actions aligned with the expected behavior. 
\end{description}

After all inputs are passed through independent encoding layers, a CNN for the game map and identity layers for game state and behavior vectors, they are concatenated into a shared feature representation. The architecture consists of two fully connected hidden layers with 512 neurons and GLU activations. This part of the architecture is the same for the actor and critic networks. Then actor network outputs discrete action logits for action selection (e.g., left, right, attack). The critic network outputs a scalar value estimate used for advantage computation, which is a crucial part of the PPO algorithm \cite{ppo_paper}. Detailed network architecture is depicted in Figure \ref{fig:network} and hyperparameters used for the PPO algorithm are provided in the Table \ref{tab:hyperparameters}. During training, all agents shared the same policy and value networks, but each was conditioned on a distinct target behavior vector. This setup resulted in a diverse set of player type combinations throughout the training process. UBCL policy is trained for nearly 200 million time steps, which is equivalent to 5,500 hours of actual gameplay. The training was conducted on consumer-grade hardware equipped with an RTX 3080 GPU (12 GB), 32 GB of RAM, and an Intel i5-13600KF processor. 16 parallel environment instances were utilized during training.

\begin{figure}[t]
\centering
\includegraphics[width=0.48\textwidth]{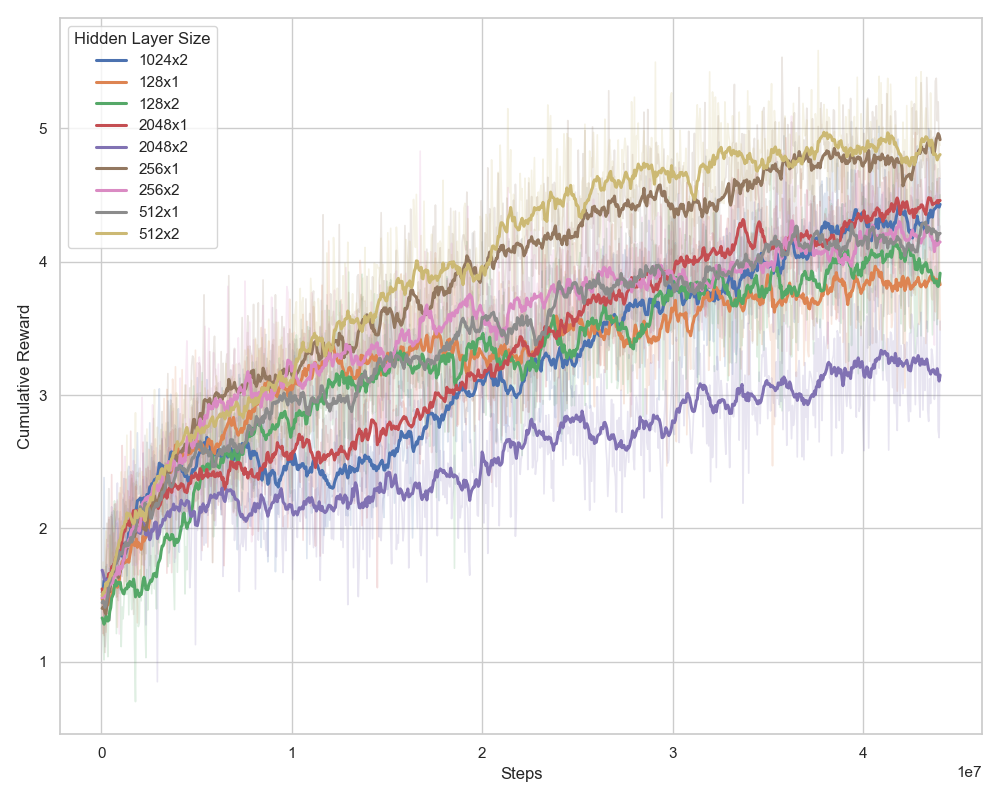}
\caption{\added{Reward convergence curves for different MLP 
configurations trained for 100M time steps under identical 
hyperparameters. The theoretical maximum cumulative reward 
per episode is $\lambda = 10$ (Eq.}~\ref{eq_reward}).}
\label{fig:mlp_comparison}
\end{figure}

\added{The distance-reduction reward (Eq.}~\ref{eq_reward}\added{) provides a dense, per-step signal that supports stable PPO 
training without requiring adjustments to default clipping or learning rate settings. Normalization by $\|b_{\mathrm{target}}\|_2$ bounds the episodic return to 
a target-independent range, limiting advantage-estimate 
variance. Additionally, the target vector is observed by 
both the actor and the critic, allowing the value function 
to condition on the current target rather than treating the 
reward as non-stationary noise.}
\added{Prior to the full training run, several MLP configurations with varying layer widths and depths were trained for 100M steps under identical hyperparameters; all exhibited stable and monotonically increasing reward curves  (Figure~}\ref{fig:mlp_comparison}\added{), and the configuration yielding the highest reward was selected for the full 200M-step training.}

\added{Among the conditioning-based frameworks reviewed in 
Section~II, CARI is the only method that provides user-facing 
control over policy inputs without requiring human demonstration 
data and is compatible with on-policy training under our 
CNN-based architecture. To provide a fair baseline, we 
implemented CARI (Configurable Agent with Reward as 
Input)}~\cite{Le_Pelletier_CARI}\added{ under identical training 
conditions: same environment, observation space, action space, 
network architecture, and training horizon. In CARI, the agent 
receives a vector of reward weights as input instead of a 
target behavior vector. At each step, the reward is computed as 
the weighted sum of the agent's current behavioral statistics. 
The reward weights are uniformly sampled per agent per episode, analogous to UBCL's target behavior vector sampling. Because CARI conditions 
on reward weights rather than behavioral targets, it provides 
directional control (``more of dimension $d$'') but does not 
support precise point targeting (``achieve exactly 
$b_d = 0.4$'').}

\begin{figure*}[!t]
\centering

\subfloat[]{\includegraphics[width=2in]{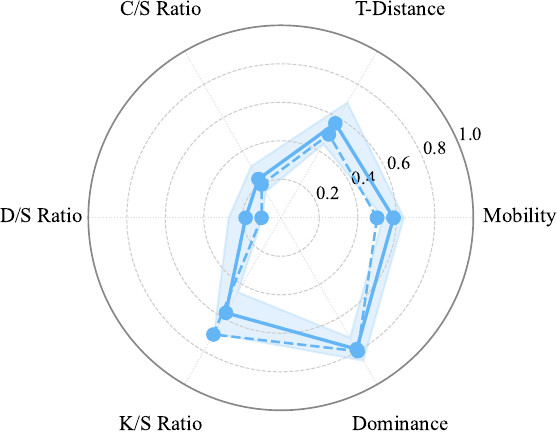}%
\label{fig_aggressive_case}}
\hfil
\subfloat[]{\includegraphics[width=2in]{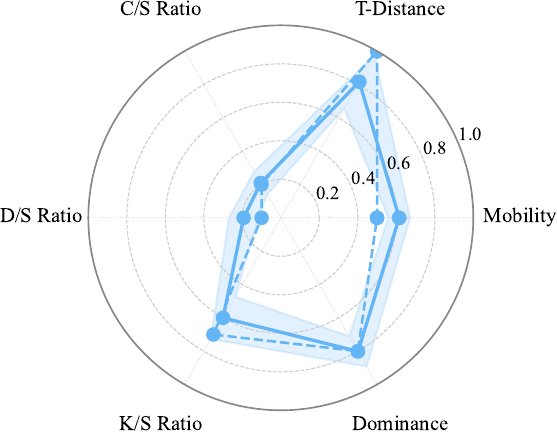}%
\label{fig_aggressive_low_coop_case}}
\hfil
\subfloat[]{\includegraphics[width=2in]{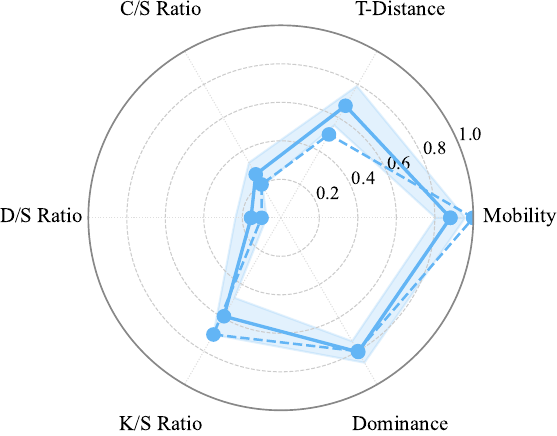}%
\label{fig_aggressive_high_mobility_case}}
\hfil

\subfloat[]{\includegraphics[width=2in]{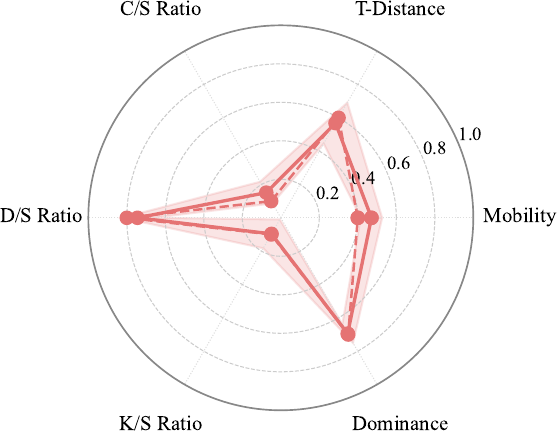}%
\label{fig_collector_case}}
\hfil
\subfloat[]{\includegraphics[width=2in]{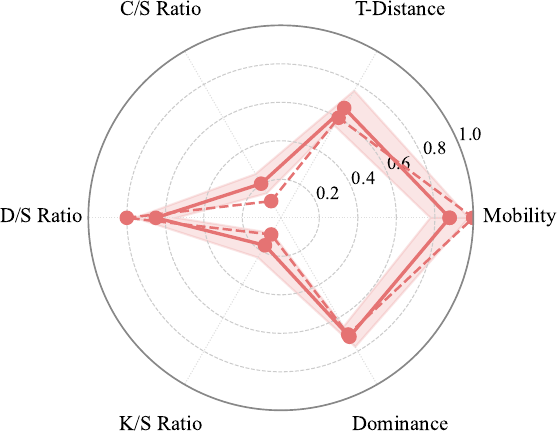}%
\label{fig_collector_high_mobility_case}}
\hfil
\subfloat[]{\includegraphics[width=2in]{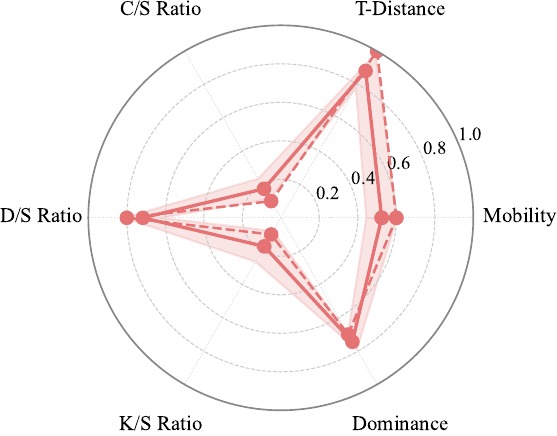}%
\label{fig_collector_low_coop_case}}
\hfil

\subfloat[]{\includegraphics[width=2in]{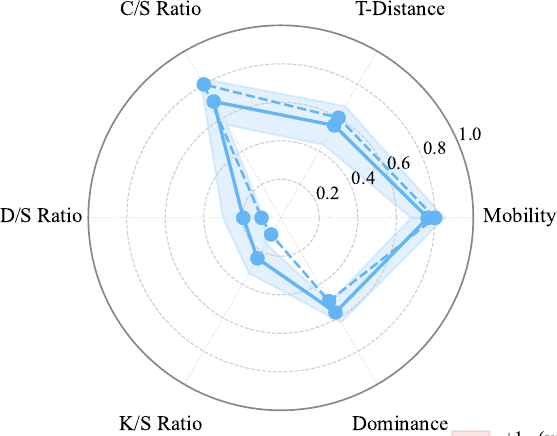}%
\label{fig_exploiter_case}}
\hfil
\subfloat[]{\includegraphics[width=2in]{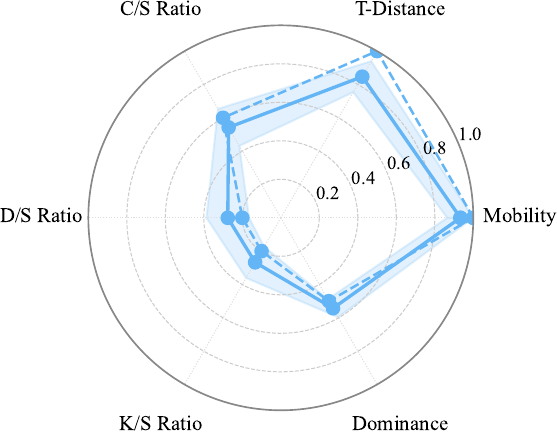}%
\label{fig_exploiter_low_coop_case}}
\hfil
\subfloat[]{\includegraphics[width=2in]{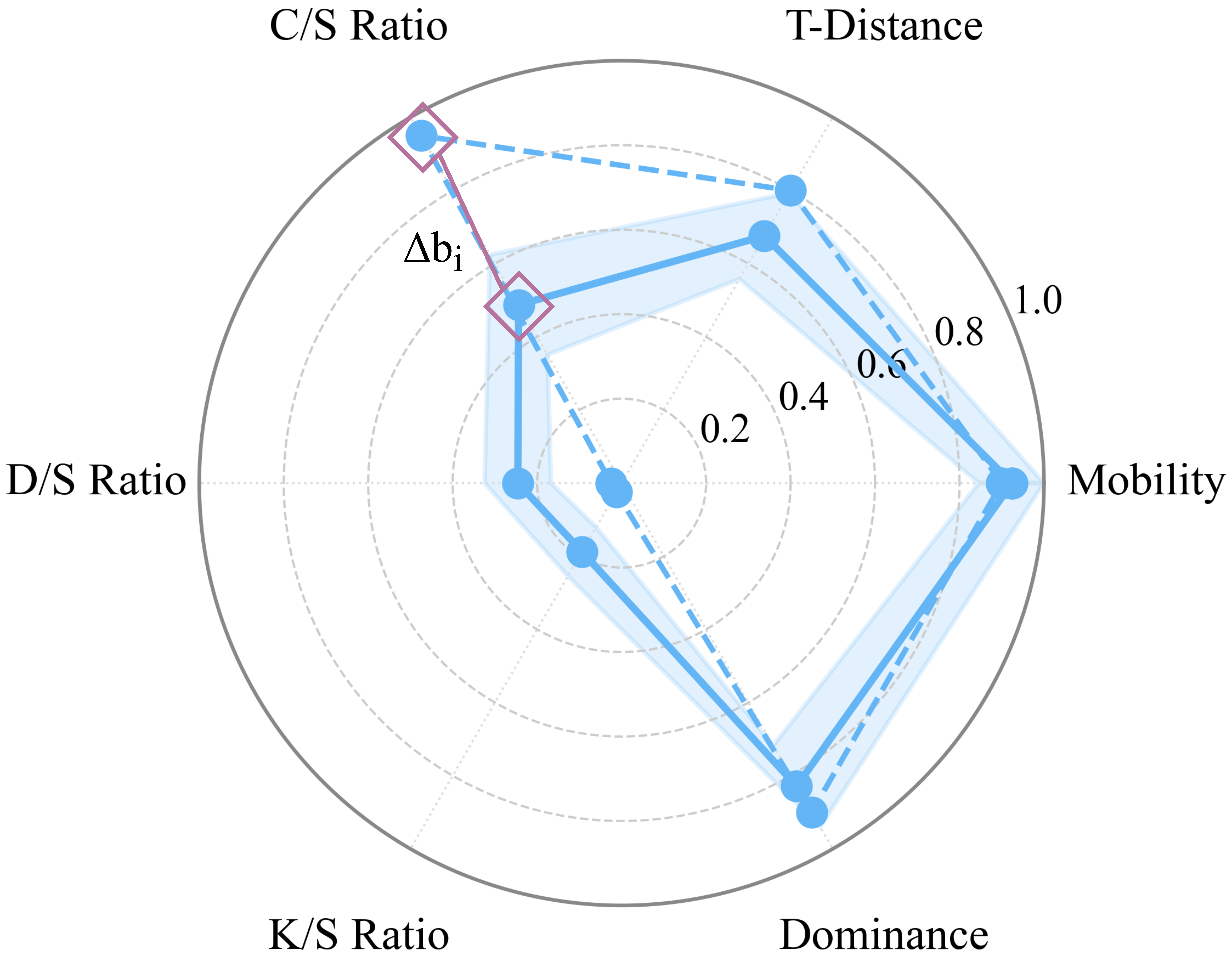}%
\label{fig_exploiter_high_dom_case}}
\hfil

\subfloat[]{\includegraphics[width=2in]{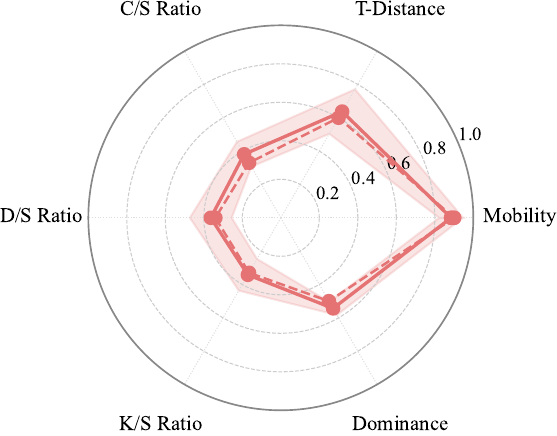}%
\label{fig_explorer_case}}
\hfil
\subfloat[]{\includegraphics[width=2in]{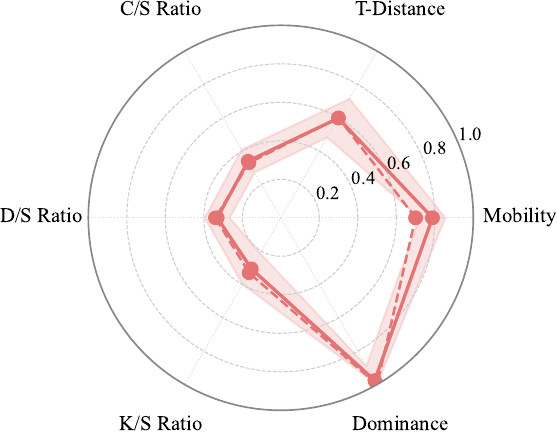}%
\label{fig_explorer_high_dom_case}}
\hfil
\subfloat[]{\includegraphics[width=2in]{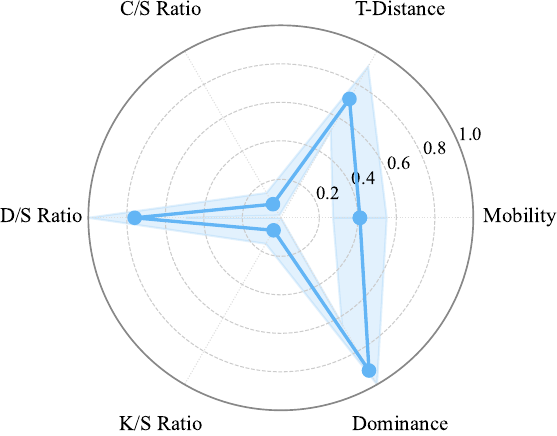}%
\label{fig_winonly}}

\caption{Learned behavior examples. Radar charts illustrate the alignment between target behavior vectors (dashed lines) and actual agent behaviors (solid lines) for four distinct profiles. Each chart is derived from 50 game episodes with a fixed target behavior vector. Shaded areas indicate the $1\sigma$ range. 
(a) Offensive agent with a high proportion of kill score.
\added{(b) Low cooperation offensive agent.}
\added{(c) High mobility offensive agent.}
\replaced{(d)}{(b)} Collector agent primarily scoring from high-risk, high-value resources.
\added{(e) High mobility collector agent.}
\added{(f) Low cooperation collector agent.}
\replaced{(g)}{(c)} Exploiter agent focusing on low-risk, widely available resources.
\added{(h) Low cooperation exploiter agent.}
\added{(i) High dominance exploiter agent. $\Delta b_i$ highlights parameters with relatively high error.}
\replaced{(j)}{(d)} Explorer agent characterized by high mobility and balanced score preferences. 
\added{(k) High dominance explorer agent.}
\replaced{(l)}{(e)} Mean behavior generated by the Win Only policy, showing only solid lines and $1\sigma$ ranges, as target behaviors are not defined for this policy.}
\label{fig:radar_charts}
\end{figure*}

\begin{table}[htbp]
\caption{Target behavior vector sample space.}
\centering
\begin{tabular}{@{}llc@{}}
\toprule
\textbf{Target Name} & \textbf{Distribution} \\ \midrule
Mobility & $\mathcal{U}[0.15, 1]$ \\
T-Distance & $\mathcal{U}[0.15, 1]$ \\
C/S ratio & $\mathcal{U}[0,1]$ \\
D/S ratio & $\mathcal{U}[0, 1 - $ C/S ratio$]$ \\
K/S ratio & 1 - C/S ratio - D/S ratio \\
Dominance & $\mathcal{U}[0, 1]$ \\
\bottomrule
\end{tabular}
\label{tab:target_distribution}
\end{table}

\begin{table}[htbp]
\caption{Hyperparameters of the PPO algorithm}
\centering
\begin{tabular}{@{}llc@{}}
\toprule
\textbf{Feature Name} & \textbf{Value} \\ \midrule
batch\_size & 1024 \\
buffer\_size & 10240 \\
learning\_rate & $3e^{-4}$ \\
learning\_rate\_schedule & linear \\
beta & $5e^{-3}$ \\
beta\_schedule & constant \\
epsilon & 0.2 \\
lambd & 0.95 \\
gamma & 0.99 \\
\bottomrule
\end{tabular}
\label{tab:hyperparameters}
\end{table}

\section{Results \deleted{and Discussion}}

\replaced{This section presents an evaluation of the trained UBCL policy in terms of behavioral accuracy, diversity, and limitations. Four complementary analyses are provided to examine how well the trained policy captures target behaviors, how diverse the resulting play styles are, where the model struggles to reproduce desired traits, and whether the policy pursues target-aligned behavior from the beginning of the episode or only converges toward the target near the end.}{This section presents an evaluation of the trained UBCL policy in terms of behavioral accuracy, diversity, and limitations. Three complementary analyses are provided to examine how well the trained policy captures target behaviors, how diverse the resulting play styles are, and where the model struggles to reproduce desired traits.}

\subsection{Behavioral Profiles}
Figure~\ref{fig:radar_charts} presents the average behavioral statistics of 50 gameplay sessions for several representative player types \added{and their variations}, compared with their corresponding target behavior vectors. For each player type, a fixed target behavior vector was assigned throughout the 50 sessions, while the target behaviors of other players were selected randomly. The results indicate that the UBCL policy generally achieves the expected behavioral targets across different gameplay scenarios. The variance of each behavioral parameter is also shown in the radar charts. Although the policy may occasionally fail to achieve a specific behavioral objective due to limited in-game resources and conflicting player goals \added{(Figure}~\ref{fig_exploiter_high_dom_case}), it consistently maintains the overall tendencies associated with each play style.

\added{We formalize controllability along two axes. Per-dimension \emph{Precision} is defined as 
$P_d = \frac{1}{N}\sum_{i=1}^{N} |b_{\mathrm{result}}^{(i),d} 
- b_{\mathrm{target}}^{d}|$, and per-dimension 
\emph{Consistency} as 
$C_d = \mathrm{std}_{i}(b_{\mathrm{result}}^{(i),d})$. 
In Figure}~\ref{fig:radar_charts}\added{, Precision corresponds to the 
target--mean gap and Consistency to the $\pm 1\sigma$ band. 
The within-style variations (b,\,c,\,e,\,f,\,h,\,i,\,k) show 
that the mean shifts as the target moves (Precision) while the 
band remains narrow within each variation (Consistency).}

To compare the generated play styles, a \emph{win-only} policy was trained, which directly maximizes the player’s score. This policy was trained for 200 million time steps, equivalent to the UBCL \added{and CARI} policy training duration. Figure~\ref{fig_winonly} shows the behavioral statistics produced by the \emph{win-only} policy over 50 games, where other agents followed the UBCL policy with randomly assigned target behaviors. As shown, the \emph{win-only} policy converged to a single dominant play style. The resulting behavioral statistics resemble the \emph{collector}-type behavior produced by the UBCL policy (Figure~\ref{fig_collector_case}). These findings suggest that the UBCL policy is capable of reproducing \emph{win-only}-like behaviors while also generating a broader diversity of play styles.

Overall, the trained agents exhibit behaviors that closely align with their intended profiles, effectively reproducing distinct archetypes such as \emph{offensive}, \emph{explorer}, and \emph{collector} types,  \added{as well as fine-grained configurations within each archetype.} \replaced{T}{Furthermore, t}he generated play styles are continuous and easily configurable under the UBCL policy. For instance, an offensive agent with higher mobility can be produced simply by increasing its mobility target \added{(Figure}~\ref{fig_aggressive_high_mobility_case}).

\subsection{Behavioral Distribution and Diversity}
\begin{figure}[!h]
\centering
\includegraphics[width=0.48\textwidth]{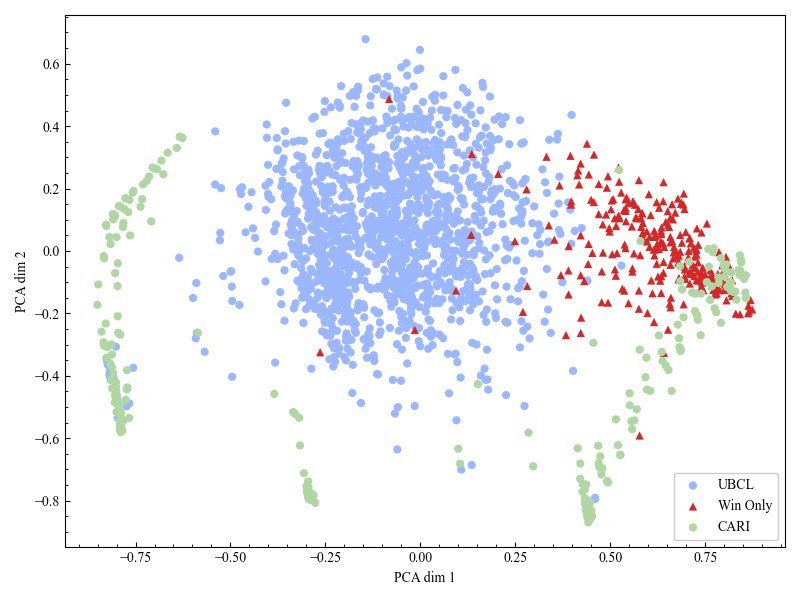}
\caption{
\deleted{Two-dimensional projection of the behavior vectors generated by the UBCL policy. A total of 1000 games were simulated, where three players were controlled by the UBCL policy conditioned on randomly sampled target behavior vectors, and one player was controlled by the \emph{win-only} policy. The resulting behavior vectors were reduced to two dimensions using PCA. Green dots represent behaviors produced by the UBCL policy, with color intensity indicating the normalized error—computed as the Euclidean distance between the target and resulting behavior vectors, divided by the magnitude of the maximum possible error vector ($\sqrt{-6}$). Blue triangles represent behavior vectors generated by the \emph{win-only} policy.}
\added{Two-dimensional PCA projection of realized 
behavior vectors. In 1000 evaluation games, one agent was 
randomly assigned to either the win-only or CARI policy, 
while the remaining agents followed the UBCL policy with 
uniformly sampled target behavior vectors to simulate various match scenarios. Blue dots: UBCL, green dots: CARI, red triangles: win-only.}}

\label{fig:2dmapping}
\end{figure}

Figure~\ref{fig:2dmapping} presents the two-dimensional \replaced{PCA projection of the six-dimensional behavior vectors produced by three policies across 1,000 evaluation games. Blue dots represent behaviors generated by the UBCL policy with uniformly sampled target vectors, green dots represent behaviors generated by the CARI policy with uniformly sampled reward weights, and red triangles represent behaviors produced by the win-only policy. The UBCL policy covers a broad region of the behavior space, reflecting its ability to reach intended targets across both interior and edge regions. The CARI policy, which provides directional control through reward weights rather than precise target specification, produces directed but edge-dominated behaviors concentrated near the boundaries of the space. The win-only policy clusters in a narrow region, as its behaviors emerge from score maximization without any behavioral intent.}{projection of six-dimensional behavior vectors obtained through Principal Component Analysis (PCA). Each point corresponds to a behavior vector generated by the UBCL policy after a game episode, with the color indicating the error magnitude relative to the randomly selected target behavior for that episode. Blue triangles denote the behavior vectors produced by the \emph{win-only} policy. As shown, the proposed UBCL model covers a wider area of the behavioral space, reflecting greater behavioral diversity. Furthermore, the play styles generated by the \emph{win-only} policy are largely encompassed within those of the UBCL policy.}

\subsection{Error Analysis per Behavioral Dimension}
\begin{figure}[!h]
  \centering
  \includegraphics[width=0.48\textwidth]{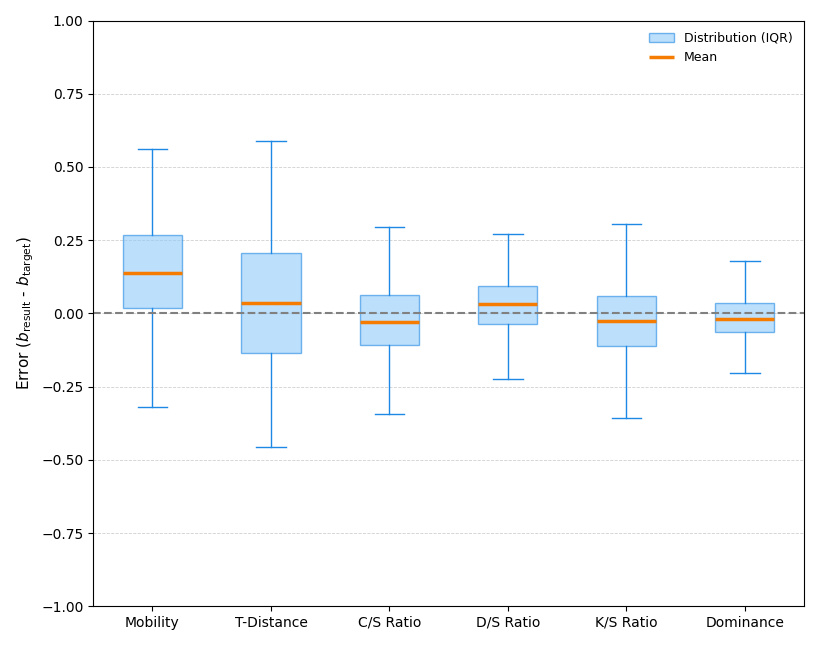}
\caption{Box-and-whisker plot illustrating the distribution of error values for each behavioral parameter across 1,250 game episodes, corresponding to 5,000 behavior vectors generated during gameplay. The error for each parameter is computed as $b_{result}-b_{target}$.}
  \label{fig:errorwhisker}
\end{figure}

Figure~\ref{fig:errorwhisker} presents the mean and variance of behavioral errors across each target dimension. The model exhibits low variance in straightforward metrics such as score-related parameters, while higher errors are observed in the \textit{T-Distance} dimension, which requires coordination with a teammate. Since teammates may pursue conflicting objectives and are conditioned by their individual target behaviors, this parameter becomes the most challenging to learn. Additionally, the mean error in \textit{Mobility} is positive and considerably high, indicating that the UBCL policy tends to produce play styles with greater mobility than expected. A plausible explanation is that the sampling interval for \textit{Mobility} includes infeasible targets, making it difficult to generate play styles with very low mobility. Consequently, the UBCL policy compensates by producing more mobile behaviors to satisfy other behavioral objectives.

\deleted{
Overall, the results demonstrate that the proposed UBCL framework preserves behavioral diversity while maintaining high fidelity to the target behavior vectors. However, the error distribution reveals specific behavioral components that could benefit from refined conditioning or additional regularization in future studies. One notable advantage of the UBCL approach is its ability to incorporate human data into the training process. By redefining the sampling intervals based on human demonstrations, more meaningful behavioral targets can be generated.
A key limitation of the UBCL framework lies in its reliance on carefully designed behavioral parameters. In this study, player behavior is described by six parameters, which inevitably constrain the representation of certain behavioral traits. For instance, motion smoothness could be an additional dimension to better characterize play styles, and accurately representing a player’s behavior might require hundreds of such parameters. Therefore, the generated play styles can be regarded as quantized approximations of highly detailed human behaviors.
}

\subsection{\added{In-Match Behavior Trajectories}}

\begin{figure*}[t]
\centering
\includegraphics[width=0.84\textwidth]{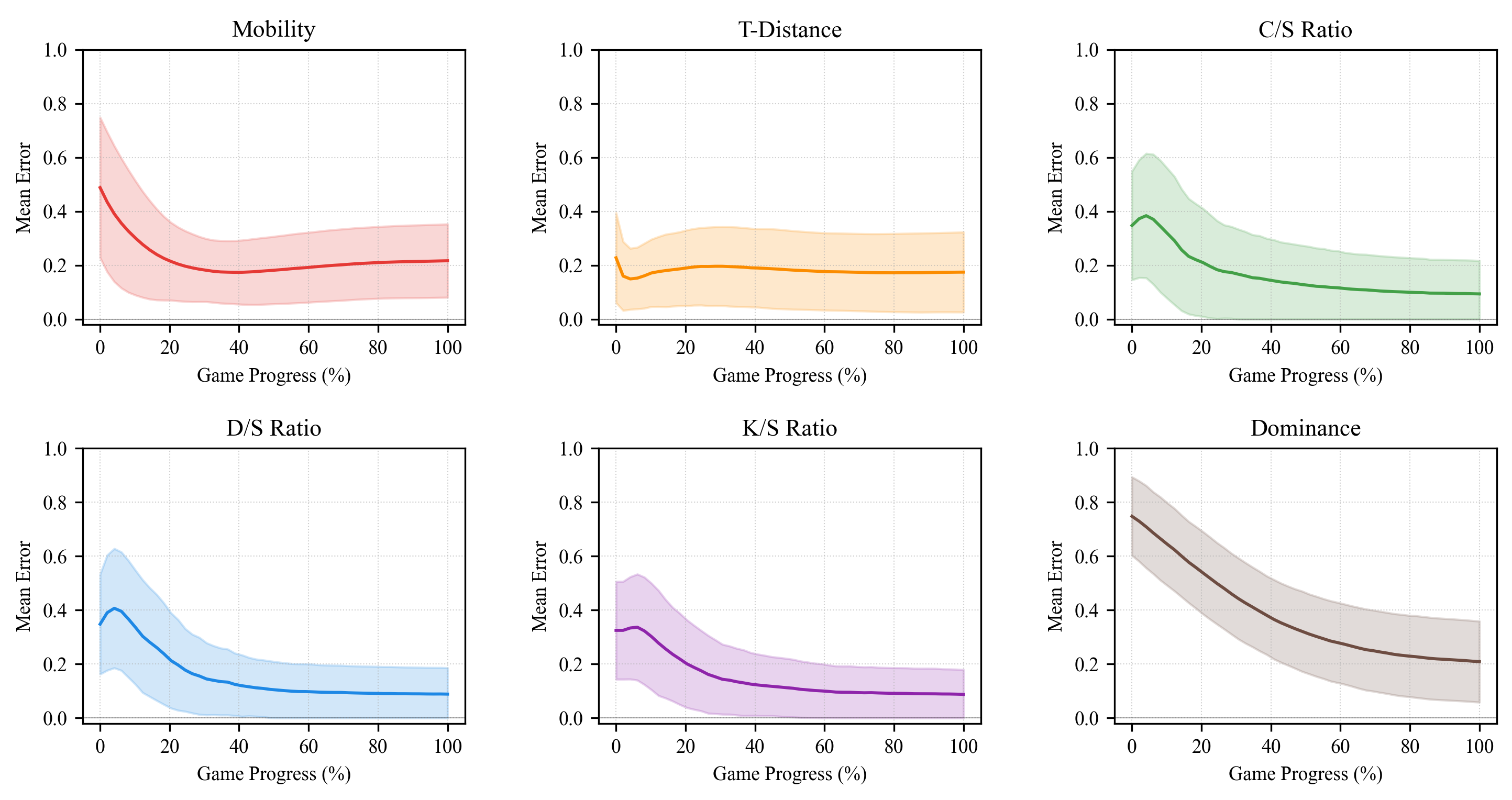}
\caption{\added{Per-dimension mean error and $\pm 1\sigma$ band 
over normalized game time, aggregated across 1,250 evaluation 
episodes.}}
\label{fig:error_trajectories}
\end{figure*}

\added{Figure}~\ref{fig:error_trajectories} \added{ shows how 
per-dimension error evolves throughout the episode. 
Ratio-based and constraint-like dimensions (C/S, D/S, K/S, Mobility, T-Distance) converge to their 
targets early and remain stable, since these metrics are updated at every step through the agent's immediate actions. Cumulative dimensions 
(Dominance) decrease in error monotonically, as 
they depend on the accumulated score that can only grow 
over time. These two convergence patterns show that the policy reaches achievable targets as early as possible and maintains them throughout the episode for ratio-based and constraint-like dimensions, while progressively approaching end-state targets for cumulative dimensions. Both behaviors emerge naturally from the same distance-reduction reward (Eq.}~\ref{eq_reward}\added{) without requiring separate objectives.}

\section{\added{Discussion}}
\added{This section examines the structural properties of the 
learned behavior space, identifies the boundary between 
individual behavior matching and team-level coordination, 
and situates UBCL relative to alternative 
diversity-generating paradigms.}

\subsection{\added{Behavior-Space Analysis}}

\begin{figure*}[t]
\centering
\includegraphics[width=0.8\textwidth]{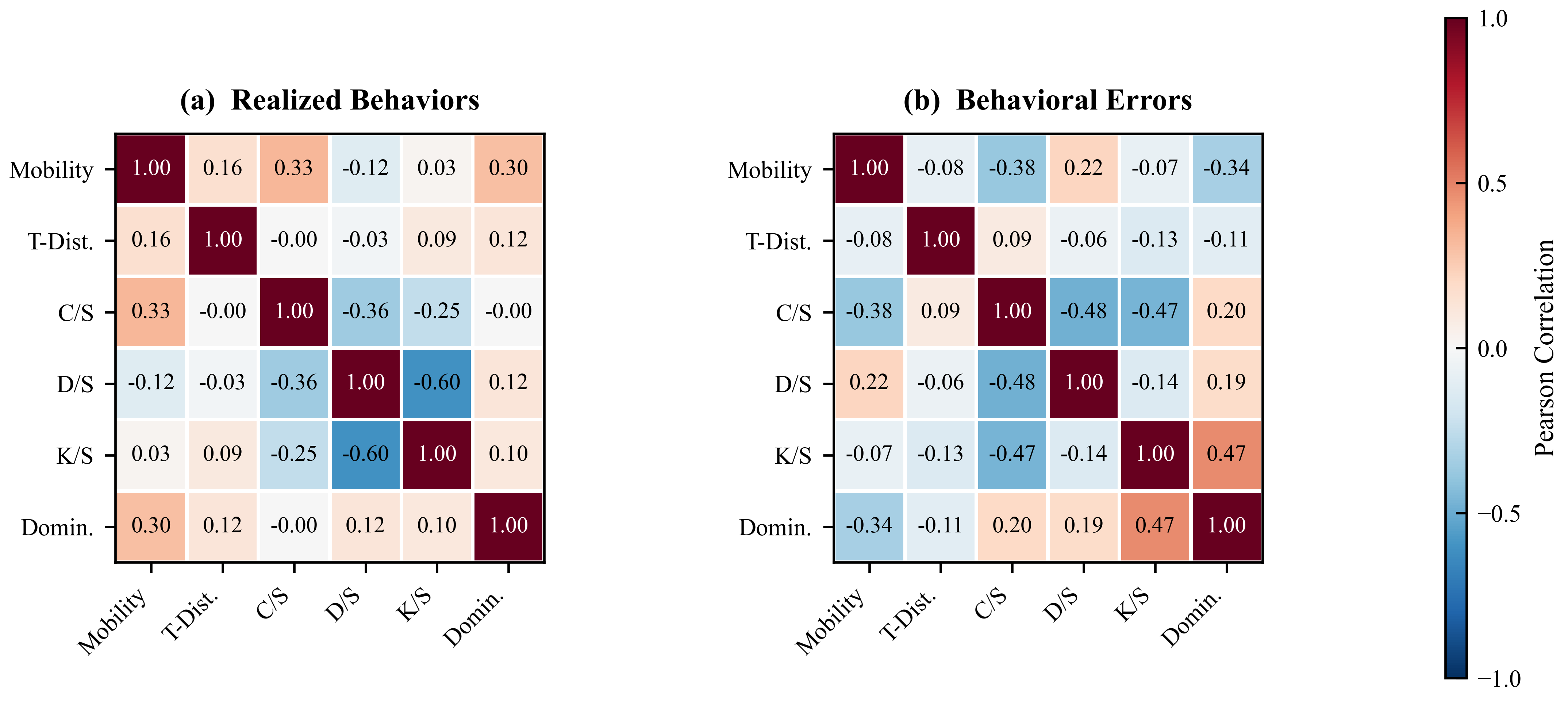}
\caption{\added{Pearson correlation matrices over 1,250 evaluation episodes. 
Left: correlations among realized behavior dimensions. 
Right: correlations among per-episode errors}
($b_{\text{result}} - b_{\text{target}}$).}
\label{fig:correlation_matrices}
\end{figure*}

\added{Figure}~\ref{fig:correlation_matrices} \added{presents 
two complementary views of the behavior space. The 
realized-behavior matrix (left) captures structural 
dependencies imposed by game mechanics. The strongest 
negative coupling is D/S--K/S ($-0.60$), which follows 
directly from the C/S\,+\,D/S\,+\,K/S\,=\,1 constraint 
rather than from a learned artifact. K/S and Dominance 
show negligible correlation ($0.10$), confirming that the 
normalized ratio and the absolute score measure capture 
distinct aspects of behavior. Mild positive correlations 
between Mobility and score-related dimensions reflect the 
fact that collecting coins and engaging opponents both 
require traversing the map; higher mobility therefore 
co-occurs naturally with higher scoring activity.}

\added{The error matrix (right) reveals emergent 
failure-mode couplings in the learned policy. K/S and 
Dominance errors co-occur ($+0.47$): when elimination-based 
scoring fails, overall score attainment suffers accordingly. 
Negative off-diagonals among ratio errors (e.g., 
C/S--D/S\,=\,$-0.48$) indicate trade-off failures in which 
the policy corrects one ratio at the cost of another, as 
imposed by the sum-to-one constraint. These patterns confirm 
that the six dimensions are not redundant proxies for a 
single latent factor, and that errors are distributed across 
dimensions rather than concentrated on one, supporting 
stable training dynamics. Certain targets fall outside the 
feasible region due to resource scarcity and competition 
from other agents (e.g., 
Figure~}\ref{fig_exploiter_high_dom_case}\added{); the policy nevertheless receives informative 
reward signal from actions directed toward such targets, 
confirming that infeasible targets contribute to learning 
rather than destabilizing training.}

\begin{figure}[t]
\centering
\includegraphics[width=0.48\textwidth]{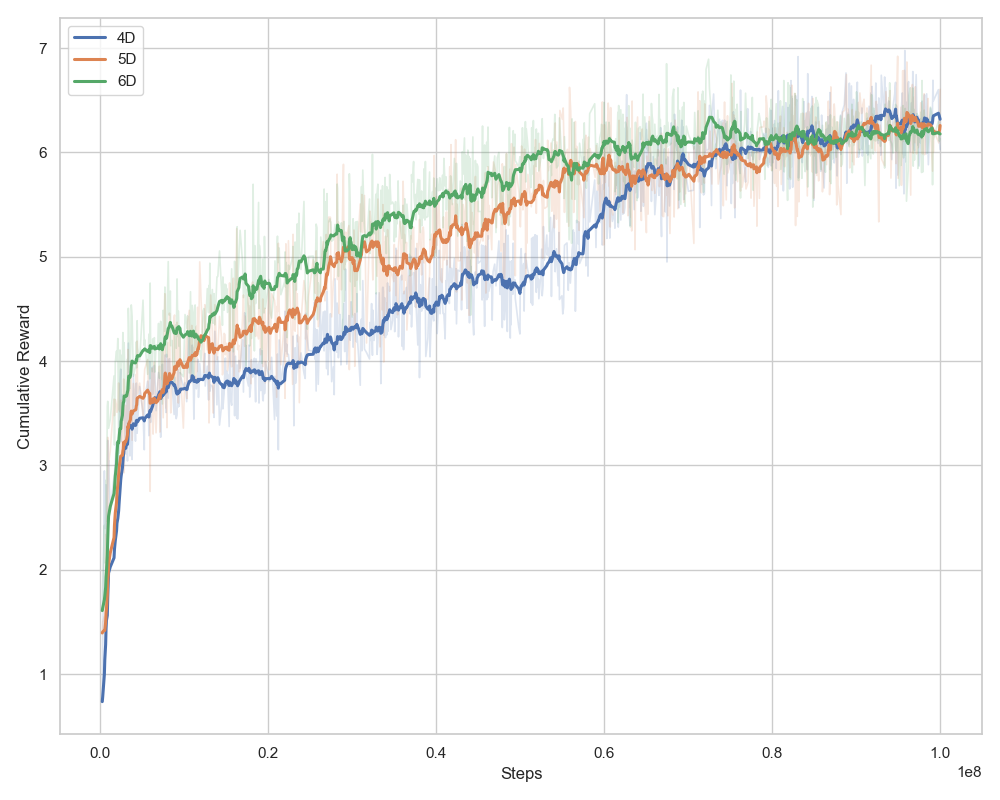}
\caption{\added{Reward convergence curves for 4, 5, and 6-dimensional behavior vector configurations trained  under identical conditions.}}
\label{fig:dim_comparison}
\end{figure}

\added{To assess the sensitivity of the framework to the 
dimensionality of the behavior vector,we trained policies with 4, 5, and 6 behavioral dimensions under identical conditions: the 4-dimensional configuration includes C/S, D/S, K/S, and Dominance; the 5-dimensional configuration adds T-Distance; and the full 6-dimensional configuration includes Mobility. Figure}~\ref{fig:dim_comparison} \added{shows that all 
three configurations exhibit comparable convergence 
dynamics, suggesting that the distance-reduction reward 
scales gracefully within the tested range. However, as the 
number of dimensions grows, the volume of the target space 
expands exponentially, which may eventually require longer 
training horizons or curriculum-based sampling strategies 
to maintain coverage.}

\added{A key limitation of the framework lies in its reliance 
on manually designed behavioral parameters. In this study, 
player behavior is described by six dimensions, which 
inevitably constrain the representable space of play styles. 
For instance, motion smoothness or tactical decision timing 
could serve as additional dimensions, and faithfully 
capturing a player's full behavioral profile might require 
a substantially larger set of parameters. Consequently, the 
generated play styles can be regarded as discretized 
approximations of highly detailed human behaviors.}

\begin{table*}[t]
\centering
\caption{\added{Comparison of diversity-generating paradigms across five criteria.}}
\label{tab:paradigm_comparison}
\begin{tabular}{@{}l c c c c c@{}}
\toprule
\textbf{Method} 
& \textbf{\shortstack{User\\Controllability}} 
& \textbf{\shortstack{Diversity\\Mechanism}} 
& \textbf{\shortstack{Policy\\Count}} 
& \textbf{\shortstack{Human Data\\Required}} 
& \textbf{\shortstack{Coverage\\Type}} \\
\midrule
DIAYN~\cite{eysenbach2019diayn}
& \shortstack{None; latent skill\\sampled automatically}
& \shortstack{MI maximization\\(skill $\leftrightarrow$ state)}
& \shortstack{Single\\(skill-conditioned)}
& No
& \shortstack{Emergent;\\no coverage guarantee} \\
\addlinespace
MAP-Elites~\cite{mouret2015mapelites}
& \shortstack{None; archive filled\\by evolutionary search}
& \shortstack{Elite archive across\\behavior descriptor bins}
& \shortstack{Multiple\\(one per bin)}
& No
& \shortstack{Discretized grid;\\resolution-limited} \\
\addlinespace
CARI~\cite{Le_Pelletier_CARI}
& \shortstack{Indirect; reward weights\\(directional control)}
& \shortstack{Varying reward\\coefficients}
& \shortstack{Single\\(weight-conditioned)}
& No
& \shortstack{Edge-dominated;\\sparse interior} \\
\addlinespace
\textbf{UBCL (ours)}
& \shortstack{\textbf{Direct; user-specified}\\\textbf{behavior vector (point targeting)}}
& \shortstack{Uniform sampling\\of target vectors}
& \shortstack{Single\\(target-conditioned)}
& No
& \shortstack{Interior + edges\\of behavior space} \\
\bottomrule
\end{tabular}
\end{table*}

\subsection{\added{Individual Behavior Matching vs. Team Coordination}}

\added{T-Distance exhibits consistently higher error than 
the other five dimensions 
(Figure}~\ref{fig:errorwhisker}\added{). Unlike score-based 
metrics, which depend only on the agent's own actions, 
T-Distance requires sustained spatial coordination with a 
teammate whose behavior is governed by an independent 
target vector. The per-agent distance-reduction reward 
averages out these interaction patterns into a single 
scalar, providing insufficient signal for joint 
spatiotemporal control. This distinction between 
individual behavior matching and team-level coordination 
is, to our knowledge, not explicitly drawn in prior 
conditioning-based frameworks. Addressing it may require 
interaction-aware objectives such as graph-based 
inter-agent distances or mutual-information-based 
coordination metrics.}

\subsection{\added{Comparison with Related Paradigms}}

\added{Table}~\ref{tab:paradigm_comparison}\added{ situates UBCL relative to three diversity-generating paradigms. DIAYN and 
MAP-Elites discover diverse behaviors automatically but do 
not accept user-specified targets, precluding direct 
controllability comparisons. CARI provides directional 
control via reward weights but converges to edge behaviors 
in the PCA projection (Figure}~\ref{fig:2dmapping}\added{), 
consistent with its weight-based formulation favoring 
extreme trade-offs. UBCL's point-targeting formulation covers the interior of the behavior space homogeneously while still reaching edge regions, whereas CARI's weight-based 
formulation concentrates on extremes.}

\added{A further practical advantage of UBCL is its ability to 
optionally incorporate human data without architectural 
changes. By redefining the sampling intervals of target 
behavior vectors based on human gameplay statistics, the 
training distribution can be concentrated on the region of 
the behavior space that real players occupy, potentially 
improving precision in that region while retaining coverage 
of the full space.}

\section{Conclusion}
This paper presented the Uniform Behavior Conditioned Learning (UBCL) framework, a reinforcement learning–based approach for generating diverse and controllable play styles within a multi-player game environment. Unlike traditional methods that rely on human demonstration data or multiple specialized policies, UBCL learns to produce distinct play styles directly from reward feedback under a single shared policy. By conditioning the learning process on target behavior vectors—including even infeasible or conflicting ones—the framework encourages exploration across the behavioral space, resulting in a richer spectrum of emergent play styles. The experiments demonstrate that UBCL effectively reproduces multiple behavioral patterns, such as \emph{offensive}, \emph{explorer}, \deleted{and} \emph{collector} types\added{, and their variations}, \replaced{with highest fidelity on score-based dimensions and moderate degradation on team coordination and mobility.}{while maintaining consistent controllability and balance among them.}

Furthermore, the utilized behavior vectors are derived from game metadata and automatically computed for each player, allowing the UBCL policy to directly represent human-player behavioral statistics. \added{Concretely, the framework can be used to test whether a map layout disadvantages specific play styles, replace disconnected players with behavior-matched substitutes derived from recent game metadata, generate stress-test opponents to surface exploits, and perform 
balance diagnostics by varying a single behavioral 
dimension while holding the others fixed.}

However, certain behavioral dimensions related to cooperation and long-term strategy remain challenging due to the short-horizon reward formulation. \replaced{Future work includes interaction-aware coordination objectives for team-level control, learned latent behavior spaces via autoencoders to replace manual metric design, validation in 
continuous-action environments with alternative distance formulations, and a comprehensive evaluation comprising cross-match metrics, teammate--opponent interaction analysis, and a dedicated human study to validate the framework's ability to reproduce individual player styles.}{Future work will focus on extending this framework to more complex multiplayer environments that demand temporal coordination and long-term planning.}

\section*{Acknowledgments}
The authors acknowledge the use of the ChatGPT-5.1 language model solely for improving the clarity and readability of the English text in this manuscript. All research ideas, methodological contributions, experiments, analyses, and conclusions presented in this work are entirely the authors’ own.

\bibliographystyle{IEEEtran}
\bibliography{references}




 

\end{document}